\documentclass[journal]{IEEEtran}
\ifCLASSINFOpdf
\else
\fi
\usepackage{multirow}
\usepackage{graphicx}
\usepackage{amsmath} 
\usepackage{url} 
\usepackage{amssymb}
\usepackage{algorithm}
\usepackage{algorithmic}
\usepackage{booktabs}
\begin{document}
%
\title{GaitKD: A Universal Decoupled Distillation Framework for Efficient Gait Recognition}
%
%
%

\author{Yuqi~Li\textsuperscript{$\dagger$},~\IEEEmembership{Member,~IEEE,}
        Qian~Zhou\textsuperscript{$\dagger$},
        Huiran~Duan,
        Jingjie~Wang.
        Shunli~Zhang.
        Chuanguang~Yang,\\
        Guoying~Zhao,~\IEEEmembership{Fellow,~IEEE,}
        and~Yingli~Tian\textsuperscript{*},~\IEEEmembership{Fellow,~IEEE,}
\thanks{\textsuperscript{$\dagger$}Yuqi~Li and Qian~Zhou contributed equally to this work.}
\thanks{\textsuperscript{*}Corresponding author: Yingli~Tian.}
\thanks{Y. Li, H. Duan, and Y. Tian are with The City University of New York (e-mail: yli067@citymail.cuny.edu; hduan@gradcenter.cuny.edu; ytian@ccny.cuny.edu).}
\thanks{Q. Zhou is with Wuhan University (e-mail: zhouqian@whu.edu.cn).}
\thanks{J. Wang and S. Zhang with Beijing Jiaotong University (e-mail: 23111492@bjtu.edu.cn; slzhang@bjtu.edu.cn).}
\thanks{C. Yang is with the Institute of Computing Technology, Chinese Academy of Sciences (e-mail: yangchuanguang@ict.ac.cn).}
\thanks{G. Zhao is with the University of Oulu (e-mail: guoying.zhao@oulu.fi).}
}

%
%

\markboth{Journal of \LaTeX\ Class Files,~Vol.~14, No.~8, August~2015}%
{Shell \MakeLowercase{\textit{et al.}}: Bare Demo of IEEEtran.cls for IEEE Journals}
%



\maketitle

\begin{abstract}
Gait recognition is an attractive biometric modality for long-range and contact-free identification, but high-performing gait models often rely on deep and computationally expensive architectures that are difficult to deploy in practice. Knowledge distillation (KD) offers a natural way to transfer knowledge from a powerful teacher to an efficient student; however, standard KD is often less effective for part-structured gait models, where supervision is formed from both part-wise classification logits and part-wise retrieval embeddings. In this paper, we propose \textit{GaitKD}, a distillation framework that decouples gait knowledge transfer into two complementary components: decision-level distillation and boundary-level distillation. Specifically, GaitKD aligns the teacher and student through part-calibrated logit distillation to transfer inter-class decision relations, while preserving the teacher-induced partitioning of the embedding space through an activation-boundary objective instead of direct feature regression. With a simple aligned part-wise design, GaitKD supports heterogeneous teacher-student gait models without introducing additional inference cost. Experimental results across multiple gait recognition benchmarks and teacher-student configurations show consistent improvements over strong gait baselines. Our study demonstrates that the two transfer components are complementary, and boundary-preserving distillation provides more stable performance than direct feature regression. Source code is available at: \url{https://github.com/liyiersan/GaitKD/}.
\end{abstract}


\begin{figure*}[thpb]
    \centering
    \includegraphics[scale=0.6]{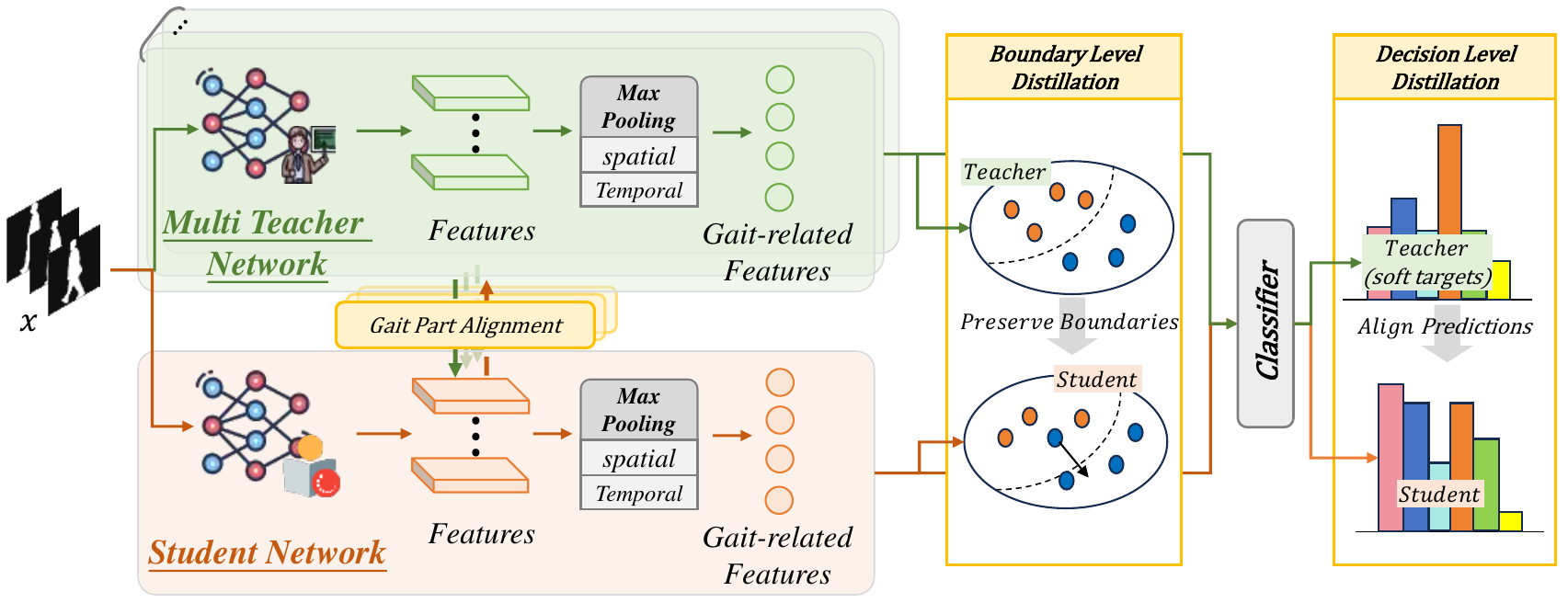}
    \caption{Overview of \textbf{GaitKD} for heterogeneous gait distillation.
    Given an input gait sequence, both a frozen teacher and an efficient student extract part-structured gait representations.
    To enable distillation across heterogeneous models, their part-wise outputs are aligned in a shared part-wise space through a \emph{gait part alignment} module.
    On top of this shared space, GaitKD performs two complementary knowledge transfers:
    \textbf{decision-level distillation}, which aligns part-wise class distributions to transfer decision relations, and
    \textbf{boundary-level distillation}, which preserves teacher-induced discriminative boundaries in the embedding space instead of directly regressing feature values.
    The two objectives are optimized jointly with the student base task loss during training.
    At inference time, only the student is used, with no additional overhead.
    For clarity, the \emph{base task supervision} (e.g., identification and metric learning losses) is omitted from the figure.}
    \vspace{-1em}
    \label{fig1}
\end{figure*}
\section{INTRODUCTION}
Gait recognition identifies individuals by their walking patterns and has become an attractive biometric modality due to its long-range and contact-free nature \cite{sepas2022deep,10735362}. Despite substantial progress, practical deployment still faces a persistent trade-off: high-performing gait models increasingly rely on deep architectures~\cite{fan2025opengait,11175576}, sophisticated temporal modeling~\cite{lin2022gaitgl,fan2020gaitpart,11474593}, or richer input representations~\cite{10612818} to improve robustness under viewpoint, clothing, and carrying variations, but these gains often come with considerable computational and engineering costs. As a result, lightweight gait models remain appealing for real-world systems, yet they typically fall behind high-capacity models, especially in unconstrained settings \cite{zheng2022gait}.

Knowledge distillation (KD) offers a principled route to narrow this gap by transferring knowledge from a powerful teacher to an efficient student \cite{hinton2015distilling}. However, standard KD is not well matched to modern gait recognition pipelines. Many gait models are part-structured, producing both part-wise classification logits and part-wise retrieval embeddings~\cite{11483236}. In this setting, distilling only outputs or directly regressing features is often insufficient~\cite{11424610}, because the teacher's knowledge is distributed across two complementary aspects: inter-class decision relations in the logit space and discriminative partitioning in the embedding space \cite{park2019relational,Tian2020Contrastive}. This challenge becomes more pronounced when the teacher and student are heterogeneous, such as a high-capacity 3D CNN or transformer-based teacher versus a compact CNN student: strict feature matching can over-constrain the student, while logit-only distillation may fail to shape robust part-aware representations.

To address this challenge, we propose \textit{GaitKD}, a distillation framework for heterogeneous, part-structured gait models. GaitKD consists of two complementary transfer objectives defined on part-wise outputs. The first is decision-level distillation, which aligns softened part-wise class distributions to transfer the teacher's inter-class decision relations. The second is boundary-level distillation, which regularizes part-wise embeddings through an activation-boundary objective. Since gait recognition is ultimately retrieval-oriented, the goal of embedding transfer is not to reproduce the teacher's exact feature values, but to preserve how identities are separated in the representation space. This makes boundary-preserving transfer better suited than direct feature regression for heterogeneous teacher-student pairs. As illustrated in Fig.~\ref{fig1}, GaitKD distills part-wise logits through temperature-scaled KL divergence with part-wise calibration, and distills part-wise embeddings through an activation-boundary objective that encourages the student to preserve discriminative boundaries rather than regress feature values directly.

Moreover, GaitKD is intentionally simple: it operates on part-wise logits and embeddings that are already produced by standard gait models, requires no modification to the teacher or student backbone, and introduces no additional inference cost. This makes it easy to integrate into existing gait training pipelines for heterogeneous teacher-student settings. Empirically, our study demonstrates that combining decision-level distillation and boundary-level distillation yields consistent improvements over strong gait baselines, validating that the two transfers are complementary rather than redundant. Our contributions are summarized as follows:

\begin{list}{$\bullet$}{
  \setlength{\leftmargin}{1.2em}
  \setlength{\itemsep}{0.2em}
  \setlength{\parsep}{0pt}
  \setlength{\topsep}{0.2em}
}
  \item We identify a structural challenge in gait knowledge distillation: for part-structured gait models, the teacher's knowledge is reflected in both part-wise class relations and embedding-space discrimination, making direct feature matching less suitable for heterogeneous teacher-student pairs.
  \item We propose \textit{GaitKD}, a distillation framework that decomposes gait knowledge transfer into decision-level distillation on part-wise logits and boundary-level distillation on part-wise embeddings, enabling effective transfer without modifying the backbone or increasing inference cost.
  \item We conduct extensive experiments on gait recognition benchmarks and heterogeneous teacher-student settings, showing that the two transfer branches are complementary and that boundary-preserving embedding transfer is more stable than direct feature regression.
\end{list}

\section{Related Work}
\subsection{Gait Recognition}
Gait recognition identifies individuals from their walking patterns and is attractive because it is non-intrusive and effective at long distances. Early work mainly relied on handcrafted representations, among which the Gait Energy Image (GEI) became a representative appearance-based descriptor by averaging silhouette sequences into a compact template \cite{han2005individual}. With the introduction of benchmark datasets such as CASIA-B \cite{yu2006framework} and OU-MVLP \cite{takemura2018multi}, the field gradually shifted toward learning-based methods that better address cross-view and cross-condition variation.

Recent progress in gait recognition has been dominated by deep models using silhouettes as input. GaitSet \cite{chao2019gaitset} reformulated gait recognition as a set-based learning problem and achieved view-invariant representation by relaxing strict temporal ordering. Later work introduced stronger spatial-temporal modeling. GaitPart \cite{fan2020gaitpart} emphasized part-level temporal cues, while GaitGL \cite{lin2022gaitgl} combined global and local representations with localized temporal aggregation to improve robustness across views. These studies show the importance of fine-grained spatial-temporal information for discriminative gait representation.

More recent research has extended gait recognition beyond conventional silhouette-only pipelines. BigGait \cite{ye2024biggait} explored self-supervised large vision models for gait representation learning. SkeletonGait and SkeletonGait++ \cite{fan2024skeletongait} incorporated structural information by combining skeleton maps with silhouettes. For challenging real-world scenarios, GaitMoE \cite{huang2024occluded} addressed occluded gait recognition with a mixture-of-experts design and introduced the OccGait benchmark. Other methods further diversified representation forms: GaitContour \cite{guo2025gaitcontour} proposed an efficient contour-pose representation, DenoisingGait \cite{jin2025denoising} suppressed identity-irrelevant appearance cues, and Gait-X \cite{wang2025gaitx} investigated generalized gait modalities from a frequency perspective.

Practicality and reproducibility have also become important themes. OpenGait \cite{fan2023opengait} provided a unified framework and proposed GaitBase, showing that careful training strategies can yield strong performance even with simple architectures. DeepGaitV2 \cite{fan2025opengait} revisited model depth and explicit temporal modeling, showing the benefit of higher-capacity architectures in realistic settings. Meanwhile, large-scale in-the-wild datasets such as GREW \cite{zhu2021gait} and Gait3D \cite{zheng2022gait} highlighted the challenges of unconstrained gait recognition and motivated more robust paradigms. Overall, existing methods reveal a clear trade-off between accuracy and efficiency, which motivates transferring knowledge from stronger models to lightweight ones.

\subsection{Knowledge Distillation}
Knowledge distillation (KD) transfers knowledge from a high-capacity teacher to a compact student model. The seminal work of Hinton et al.~\cite{hinton2015distilling} introduced logit-based distillation, where softened class distributions provide richer supervision than hard labels by encoding inter-class similarity. This idea has been widely adopted in visual recognition.

Later studies extended KD from outputs to intermediate representations. FitNets~\cite{romero2015fitnets} used intermediate hints, and attention transfer~\cite{zagoruyko2016paying} encouraged students to mimic teacher attention patterns. Other approaches focused on structural or relational knowledge, including FSP matrices~\cite{yim2017gift}, factor transfer~\cite{kim2018paraphrasing}, relational knowledge distillation (RKD)~\cite{park2019relational,yang2022cross}, and contrastive representation distillation (CRD)~\cite{Tian2020Contrastive,yang2023online}. HSAKD~\cite{yang2021hierarchical} further introduced self-supervised augmented distillation to improve student representations.

Recent work has revisited both feature- and logit-level KD. Overhaul of Feature Distillation~\cite{heo2019comprehensive} improved optimization for feature transfer, while Activation Boundary distillation~\cite{heo2019knowledge} focused on matching activation patterns rather than exact values. On the logit side, Decoupled Knowledge Distillation (DKD)~\cite{zhao2022decoupled} separated target and non-target information, and later studies refined output transfer through multilevel alignment \cite{jin2023multilevel}, logit standardization \cite{sun2024logit}, and finer-grained decomposition \cite{wei2024scaled}.

In biometric recognition, distillation has also been explored in identity-sensitive embedding tasks such as face recognition. These studies show that feature-space transfer is especially important in open-set settings and that large teacher-student gaps can make naive distillation unstable \cite{li2023rethinking,boutros2024adadistill}. Compared with general visual recognition, KD for gait recognition remains relatively underexplored. Since modern gait models often combine identity supervision with retrieval-oriented embeddings, effective gait distillation should transfer both class-relation information in logits and discriminative structure in embeddings, especially for heterogeneous teacher-student architectures. This motivates the framework studied in this work.

\addtolength{\textheight}{-3cm}   

\section{Method}
\label{sec:method}
This section presents \textit{GaitKD} and its training objective. We first introduce the overall formulation and the shared part-wise distillation space, and then present the student base objective, the two distillation branches, and the final optimization objective.

\subsection{Overview}
\label{sec:overview}
We propose \textit{GaitKD}, a distillation framework for part-structured gait models. The key idea is to define distillation on the task outputs that are common to modern gait recognition pipelines, rather than on arbitrary intermediate features. Specifically, both the teacher and the student produce part-wise logits for identity prediction and part-wise embeddings for retrieval. GaitKD performs knowledge transfer on these shared outputs through two complementary objectives.

The first objective is \emph{decision-level distillation}, which aligns the teacher and student on softened part-wise class distributions. This branch transfers the teacher's inter-class decision relations and provides richer supervision than hard labels alone. The second objective is \emph{boundary-level distillation}, which constrains the student to preserve the teacher-induced discriminative boundaries in the embedding space. Since gait recognition is ultimately retrieval-oriented, the goal of embedding transfer is not to reproduce the teacher's exact feature values, but to preserve how identities are separated in the representation space. 
This makes boundary-preserving transfer more suitable than direct feature regression for heterogeneous teacher-student pairs.

Based on this formulation, GaitKD consists of three components: 
(i) a shared part-wise distillation space that aligns teacher and student outputs, 
(ii) a decision-level branch on part-wise logits, and 
(iii) a boundary-level branch on part-wise embeddings. 
The two distillation branches are jointly optimized with the student's base gait objective during training, while inference uses only the student and introduces no additional cost.

\subsection{Shared Part-wise Distillation Space}
\label{sec:sharedspace}
Let $f_s(\cdot;\theta_s)$ and $f_t(\cdot;\theta_t)$ denote the student and teacher networks, respectively, where the teacher is frozen during training. 
Given a mini-batch $\{(X_i,y_i)\}_{i=1}^{B}$ with identity labels $y_i\in\{1,\dots,C\}$, both networks produce two types of part-wise outputs:
\begin{equation}
    f_m(X)\mapsto \bigl(Z_m(X),\,E_m(X)\bigr), \qquad m\in\{s,t\},
    \label{eq:iface_new}
\end{equation}
where $Z_m(X)\in\mathbb{R}^{B\times C\times P_m}$ denotes part-wise classification logits and $E_m(X)\in\mathbb{R}^{B\times D_m\times P_m}$ denotes part-wise embeddings. Here $P_m$ is the number of parts and $D_m$ is the embedding dimension of model $m$. The tensor axes follow $(\text{batch},\text{channel},\text{part})$. For example, $Z_m(i,c,p)$ is the logit of class $c$ for sample $i$ at part $p$, and $E_m(i,d,p)$ is the $d$-th embedding coordinate at the same part. For convenience, the $p$-th slice of a tensor $A\in\mathbb{R}^{B\times K\times P}$ is written as
\begin{equation}
    A^{(p)} := A[:,:,p]\in\mathbb{R}^{B\times K}.
\end{equation}

GaitKD performs distillation on a shared part-wise space derived from these outputs. When the teacher and student use the same part granularity, their part-wise outputs can be compared directly. When their part granularities differ, we adopt a simple parameter-free alignment by restricting both models to a common part dimension
\begin{equation}
    P \triangleq \min(P_s,P_t).
    \label{eq:Pmin_new}
\end{equation}
The aligned logits and embeddings are then written as
\begin{equation}
    \begin{aligned}
    \widetilde{Z}_s &= Z_s[:,:,1\!:\!P], \qquad & \widetilde{Z}_t &= Z_t[:,:,1\!:\!P],\\
    \widetilde{E}_s &= E_s[:,:,1\!:\!P], \qquad & \widetilde{E}_t &= E_t[:,:,1\!:\!P].
    \end{aligned}
    \label{eq:cropP_new}
\end{equation}

This shared part-wise space serves as the basis for both distillation branches. Decision-level distillation is defined on $(\widetilde{Z}_s,\widetilde{Z}_t)$, while boundary-level distillation is defined on $(\widetilde{E}_s,\widetilde{E}_t)$. By avoiding strict layer-by-layer matching and operating only on aligned task outputs, GaitKD remains simple to apply across heterogeneous gait models.

\subsection{Gait Recognition Base Objective}
\label{sec:taskloss}
This subsection specifies the \emph{base} training objective for the student, i.e., the standard gait recognition losses
without any distillation terms. The objective combines an identification loss on part-wise logits and a retrieval loss on
part-wise embeddings.

\paragraph{Part-wise identification loss}
Let $\widetilde{Z}_s^{(p)} := \widetilde{Z}_s[:,:,p]\in\mathbb{R}^{B\times C}$ denote the student logits of part $p$.
The corresponding class probability is defined by
\begin{equation}
\pi_s^{(p)}(i,c) = \frac{\exp(\widetilde{Z}_s^{(p)}(i,c))}
{\sum_{c'=1}^{C}\exp(\widetilde{Z}_s^{(p)}(i,c'))}.
\end{equation}
The part-averaged cross-entropy is
\begin{equation}
\mathcal{L}_{\mathrm{CE}}
= \frac{1}{BP}\sum_{p=1}^{P}\sum_{i=1}^{B} -\log \pi_s^{(p)}(i,y_i).
\label{eq:CE}
\end{equation}

\paragraph{Part-wise retrieval loss}
For part $p$, let $\widetilde{E}_s^{(p)}\in\mathbb{R}^{B\times D_s}$ denote the student embedding, and define its $\ell_2$-normalized form as
\begin{equation}
\bar{e}_{s,i}^{(p)}=
\frac{\widetilde{E}_s^{(p)}(i,:)}
{\|\widetilde{E}_s^{(p)}(i,:)\|_2}.
\end{equation}
We adopt a triplet loss on the normalized embeddings:
\begin{equation}
\mathcal{L}_{\mathrm{Tri}}
=
\frac{1}{BP}\sum_{p=1}^{P}\sum_{i=1}^{B}
\Bigl[
m_{\mathrm{tri}}
+d\!\left(\bar{e}_{s,i}^{(p)},\bar{e}_{s,i^{+}}^{(p)}\right)
-d\!\left(\bar{e}_{s,i}^{(p)},\bar{e}_{s,i^{-}}^{(p)}\right)
\Bigr]_+,
\label{eq:tri}
\end{equation}
where $[\cdot]_+=\max(\cdot,0)$.

\paragraph{Base objective}
The student base objective is
\begin{equation}
\mathcal{L}_{\mathrm{task}}
=\lambda_{\mathrm{CE}}\mathcal{L}_{\mathrm{CE}}+\lambda_{\mathrm{Tri}}\mathcal{L}_{\mathrm{Tri}}.
\label{eq:task}
\end{equation}
This base objective establishes the standard identification and retrieval supervision for learning part-structured gait representations.
In the proposed framework, all distillation terms are introduced as auxiliary regularizers on top of $\mathcal{L}_{\mathrm{task}}$,
and the teacher contributes only through its outputs (logits and embeddings) while remaining frozen.

\subsection{Decision Level Distillation: Part Calibrated Logit Transfer}
\label{sec:logit}

Decision level distillation aligns the teacher and student in the logit space by matching \emph{part-wise} softened class distributions.
Given temperature $T>0$ and a logit scaling factor $\alpha>0$, the softened distributions for part $p$ are:
\begin{equation}
\label{eq:soft}
q_m^{(p)}(i,c)
=\mathrm{softmax}\!\left(\frac{\alpha\,\widetilde{Z}_m^{(p)}(i,c)}{T}\right),
\qquad m\in\{s,t\}.
\end{equation}
The part-calibrated logit distillation loss is then formulated as a part-normalized KL divergence:
\begin{align}
\mathcal{L}_{\mathrm{KL}}
&=\frac{T^2}{BP}\sum_{p=1}^{P}\sum_{i=1}^{B}
\mathrm{KL}\!\left(q_t^{(p)}(i,\cdot)\,\Vert\,q_s^{(p)}(i,\cdot)\right)
\label{eq:KL}\\
&=\frac{T^2}{BP}\sum_{p=1}^{P}\sum_{i=1}^{B}\sum_{c=1}^{C}
q_t^{(p)}(i,c)\log\frac{q_t^{(p)}(i,c)}{q_s^{(p)}(i,c)}.
\nonumber
\end{align}
The normalization by $P$ enforces \emph{part calibration}, i.e., each part contributes equally to the distillation signal,
while the $T^2$ factor compensates for gradient attenuation induced by temperature scaling.

\paragraph{Gradient form}
Let $\ell_{i,p}$ denote the KL term for sample $i$ and part $p$ in Eq.~\eqref{eq:KL}, and define
$s=\alpha\,\widetilde{Z}_s^{(p)}(i,:)$. The gradient w.r.t.\ $s_c$ satisfies
\begin{equation}
\label{eq:gradKL}
\begin{aligned}
\frac{\partial \ell_{i,p}}{\partial s_c}
&=\frac{1}{T}\Bigl(q_s^{(p)}(i,c)-q_t^{(p)}(i,c)\Bigr),\\
\frac{\partial \mathcal{L}_{\mathrm{KL}}}{\partial \widetilde{Z}_s^{(p)}(i,c)}
&=\frac{\alpha T}{BP}\Bigl(q_s^{(p)}(i,c)-q_t^{(p)}(i,c)\Bigr).
\end{aligned}
\end{equation}
which makes explicit that part normalization yields a stable gradient scale under different part granularities.

\paragraph{Decoupled decision transfer}
Beyond full-distribution matching, the decision transfer can be expressed in a decoupled form that separately constrains
(i) the teacher's confidence on the ground-truth gait identity and (ii) the teacher's relative preferences among non-target identities.
For a given pair $(i,p)$ with label $y=y_i$, define the collapsed two-class distributions
\begin{equation}
\bar{q}_m^{(i,p)}
=\Bigl[q_m^{(p)}(i,y),\ \sum\nolimits_{c\neq y} q_m^{(p)}(i,c)\Bigr]\in\mathbb{R}^{2},
\qquad m\in\{s,t\}.
\label{eq:collapse}
\end{equation}
The corresponding target-vs-others transfer is
\begin{equation}
\mathcal{L}_{\mathrm{TCKD}}
=\frac{T^2}{BP}\sum_{p=1}^{P}\sum_{i=1}^{B}
\mathrm{KL}\!\left(\bar{q}_t^{(i,p)}~\Vert~\bar{q}_s^{(i,p)}\right).
\label{eq:TCKD}
\end{equation}
To distill the teacher's preference structure among non-target classes, define masked-softmax distributions by excluding the
ground-truth class with a large constant $\gamma$:
\begin{equation}
q_{m,\neg y}^{(p)}(i,c)
=\mathrm{softmax}\!\left(\frac{\alpha\,\widetilde{Z}_m^{(p)}(i,c)}{T}-\gamma\,\mathbf{1}[c=y]\right),
\label{eq:masksoft}
\end{equation}
where $\gamma\gg 1$,and the non-target transfer is
\begin{equation}
\mathcal{L}_{\mathrm{NCKD}}
=\frac{T^2}{BP}\sum_{p=1}^{P}\sum_{i=1}^{B}
\mathrm{KL}\!\left(q_{t,\neg y}^{(p)}(i,\cdot)~\Vert~q_{s,\neg y}^{(p)}(i,\cdot)\right).
\label{eq:NCKD}
\end{equation}
The decoupled decision-level objective is
\begin{equation}
\mathcal{L}_{\mathrm{DKD}}=\alpha_d\,\mathcal{L}_{\mathrm{TCKD}}+\beta_d\,\mathcal{L}_{\mathrm{NCKD}}.
\label{eq:DKD}
\end{equation}
For notational convenience, the decision-level distillation term in the full objective (Sec.~\ref{sec:objective}) is written as
$\mathcal{L}_{\mathrm{KL}}$.
By aligning softened part-wise class distributions, the student inherits the teacher's inter-identity relations in the logit space while maintaining a stable gradient scale across different part granularities.

\subsection{Boundary Level Distillation: Activation-Boundary Transfer}
\label{sec:boundary}

Boundary level distillation transfers the teacher's \emph{representation partitioning} in the embedding space.
Rather than enforcing point wise equality of embeddings under heterogeneous architectures, the objective preserves the
teacher-induced activation boundaries with a margin.

Let $e_{s,i,d}^{(p)} := \widetilde{E}_s^{(p)}(i,d)$ and $e_{t,i,d}^{(p)} := \widetilde{E}_t^{(p)}(i,d)$ denote the aligned
student/teacher embedding coordinates. The teacher-induced boundary indicator is defined as
\begin{equation}
\sigma_{t,i,d}^{(p)}=\mathrm{sign}\!\left(e_{t,i,d}^{(p)}\right)\in\{-1,+1\}.
\label{eq:sign}
\end{equation}

\paragraph{Margin-based boundary constraints}
A margin $m>0$ specifies a boundary band around zero. For $\sigma_{t,i,d}^{(p)}=-1$, the constraint is $e_{s,i,d}^{(p)}\le -m$;
for $\sigma_{t,i,d}^{(p)}=+1$, the constraint is $e_{s,i,d}^{(p)}\ge m$. Constraint violations are measured by
\begin{equation}
\begin{aligned}
v_{-}(e)=\mathrm{ReLU}(e+m),\\
v_{+}(e)=\mathrm{ReLU}(m-e),
\end{aligned}
\label{eq:viol}
\end{equation}
where $\mathrm{ReLU}(x)=\max(x,0)$.

\paragraph{Activation-boundary loss}
The per-coordinate boundary penalty is
\begin{equation}
\ell_{\mathrm{AB}}(e_s,e_t)
=\mathbf{1}[e_t\le 0]\cdot v_{-}(e_s)^2
+\mathbf{1}[e_t>0]\cdot v_{+}(e_s)^2,
\label{eq:ab_elem}
\end{equation}
and the boundary-level distillation objective averages over samples, parts, and embedding dimensions:
\begin{equation}
\mathcal{L}_{\mathrm{AB}}
=\frac{1}{BPD_s}\sum_{p=1}^{P}\sum_{i=1}^{B}\sum_{d=1}^{D_s}
\ell_{\mathrm{AB}}\!\left(e_{s,i,d}^{(p)},e_{t,i,d}^{(p)}\right).
\label{eq:AB}
\end{equation}

\paragraph{Vectorized expression}
Let $E_s^{(p)}:=\widetilde{E}_s[:,:,p]\in\mathbb{R}^{B\times D_s}$ and define masks
$M_-^{(p)}=\mathbf{1}[\widetilde{E}_t^{(p)}\le 0]$ and $M_+^{(p)}=\mathbf{1}[\widetilde{E}_t^{(p)}>0]$.
Then Eq.~\eqref{eq:AB} admits the vectorized form
\begin{equation}
\label{eq:AB_vec}
\begin{aligned}
\mathcal{L}_{\mathrm{AB}}
=\frac{1}{BPD_s}\sum_{p=1}^{P}\Bigl(
&\bigl\| M_-^{(p)}\odot \mathrm{ReLU}(E_s^{(p)}+m)\bigr\|_F^2 \\
+~&\bigl\| M_+^{(p)}\odot \mathrm{ReLU}(m-E_s^{(p)})\bigr\|_F^2
\Bigr),
\end{aligned}
\end{equation}
where $\odot$ denotes element-wise multiplication.

\paragraph{Multi-layer boundary distillation}
If a set of layers $\ell\in\{1,\dots,L\}$ exposes aligned part-structured embeddings $\{E_s^\ell,E_t^\ell\}$,
the boundary objective is aggregated as
\begin{equation}
\mathcal{L}_{\mathrm{AB}}^{\mathrm{multi}}
=\sum_{\ell=1}^{L} w_\ell\,\mathcal{L}_{\mathrm{AB}}(E_s^\ell,E_t^\ell),
\qquad
w_\ell\ge 0,\ \ \sum_{\ell=1}^{L} w_\ell=1.
\label{eq:AB_multi}
\end{equation}

\subsection{Multi-Teacher Distillation}
\label{sec:multiteacher}

GaitKD naturally supports multiple heterogeneous teachers $\{f_{t_k}\}_{k=1}^{K}$ (e.g., different backbones or modalities).
Each teacher provides $(Z_{t_k},E_{t_k})$.
After part alignment (Eq.~\eqref{eq:cropP_new}), we aggregate teachers at the distribution level.

\paragraph{Logit distribution ensemble}
Let $q_{t_k}^{(p)}$ be the softened distribution of teacher $k$ at part $p$.
We form an ensemble teacher distribution:
\begin{equation}
\begin{aligned}
q_{T}^{(p)}(i,c)=\sum_{k=1}^{K} w_k(i,p)\, q_{t_k}^{(p)}(i,c),\\
\qquad \sum_{k=1}^{K}w_k(i,p)=1,~~w_k(i,p)\ge 0,
\end{aligned}
\label{eq:teacher_ens}
\end{equation}
and replace $q_t^{(p)}$ by $q_T^{(p)}$ in Eq.~\eqref{eq:KL}.

\paragraph{Confidence-based weights}
Weights can be uniform $w_k=\tfrac{1}{K}$, or set by teacher confidence.
For example, using entropy $H(q)= -\sum_c q(c)\log q(c)$:
\begin{equation}
w_k(i,p)=\frac{\exp\!\left(-\tau\,H(q_{t_k}^{(p)}(i,\cdot))\right)}{\sum_{k'}\exp\!\left(-\tau\,H(q_{t_{k'}}^{(p)}(i,\cdot))\right)},
\label{eq:entropy_weight}
\end{equation}
where $\tau>0$ controls sharpness.

\paragraph{Embedding boundary aggregation}
For boundary distillation, we can distill from the strongest teacher only (max-confidence teacher),
or aggregate teacher signs by majority vote:
\begin{equation}
\sigma_T^{(p)}(i,d)=\mathrm{sign}\!\left(\sum_{k=1}^{K} w_k(i,p)\,\mathrm{sign}(e_{t_k,i,d}^{(p)})\right),
\label{eq:sign_vote}
\end{equation}
and use $\sigma_T$ to gate Eq.~\eqref{eq:ab_elem}.

\subsection{Unified Objective}
\label{sec:objective}

GaitKD combines decision-level and boundary-level transfers:
\begin{equation}
\mathcal{L}_{\mathrm{KD}}=\lambda_{\mathrm{logit}}\mathcal{L}_{\mathrm{KL}}+\lambda_{\mathrm{bound}}\mathcal{L}_{\mathrm{AB}}.
\label{eq:KD}
\end{equation}
The full training objective is
\begin{equation}
\label{eq:total}
\begin{aligned}
\mathcal{L}
&=\mathcal{L}_{\mathrm{task}}+\mathcal{L}_{\mathrm{KD}} \\
&=\lambda_{\mathrm{CE}}\mathcal{L}_{\mathrm{CE}}
+\lambda_{\mathrm{Tri}}\mathcal{L}_{\mathrm{Tri}}
+\lambda_{\mathrm{logit}}\mathcal{L}_{\mathrm{KL}} \\
&\quad +\lambda_{\mathrm{bound}}\mathcal{L}_{\mathrm{AB}}.
\end{aligned}
\end{equation}
During training, $\mathcal{L}_{\mathrm{task}}$ enforces identity discrimination and retrieval structure on the student representations,
while $\mathcal{L}_{\mathrm{KD}}$ transfers complementary teacher knowledge at the decision level (logits) and at the boundary level (embeddings).
The teacher remains frozen and affects learning solely through $(\widetilde{Z}_t,\widetilde{E}_t)$, so no additional computation is introduced at inference.

\paragraph{Hyperparameters}
GaitKD introduces temperature $T$ and boundary margin $m$, plus weights
$\lambda_{\mathrm{logit}}$ and $\lambda_{\mathrm{bound}}$ (and optionally $\alpha$).
In practice, $T$ controls the softness of decision geometry transfer (Eq.~\eqref{eq:soft}),
while $m$ controls the strictness of boundary preservation (Eq.~\eqref{eq:ab_elem}).




\section{Experiments}
\label{sec:exp}

We evaluate the proposed GaitKD framework on three gait recognition benchmarks: Gait3D~\cite{zheng2022gait}, CCPG~\cite{li2023depth}, and SUSTech1K~\cite{shen2023lidargait}. 
These datasets together cover large-scale in-the-wild gait recognition, cloth-changing scenarios, and outdoor 3D LiDAR-based gait recognition. 
All experiments follow a deployment-oriented distillation setting, where high-capacity teachers are available only during training and an efficient student is used for inference. 
Accordingly, all reported results are obtained using the \emph{student alone} at test time. 
Teachers are frozen during training and discarded after training, so GaitKD introduces no additional inference-time overhead.

Our experiments are organized around three questions:
(i) whether GaitKD improves a strong lightweight student under different teacher-student settings,
(ii) whether multiple teachers provide complementary supervision, and
(iii) how the decision-level and boundary-level components contribute through ablation studies.

\subsection{Benchmarks and Metrics}
\label{sec:benchmarks}

\paragraph{Benchmarks}
We use three complementary benchmarks. 
Gait3D~\cite{zheng2022gait} is a large-scale in-the-wild benchmark with substantial variations in viewpoint, background, and walking conditions. 
CCPG~\cite{li2023depth} focuses on cloth-changing gait recognition and evaluates robustness under appearance changes. 
SUSTech1K~\cite{shen2023lidargait} is an outdoor LiDAR-based gait benchmark with diverse challenge factors, including carrying, clothing, occlusion, and nighttime conditions. 
Evaluating on these datasets allows us to assess GaitKD across both conventional image/silhouette-based gait recognition and 3D LiDAR-based gait recognition settings.

\paragraph{Evaluation protocols}
For each benchmark, we follow the official training/test split and evaluation protocol. 
On Gait3D, evaluation follows the standard gallery--probe retrieval setting defined by the benchmark. 
On CCPG, we follow the official cloth-changing protocol and report results on the benchmark-defined subsets used in our experiments. 
On SUSTech1K, we follow the official cross-view retrieval protocol and report both overall performance and, when applicable, attribute-wise performance under different probe conditions.

\paragraph{Metrics}
We report the official evaluation metrics of each benchmark rather than enforcing a unified metric across datasets. 
For Gait3D, we report \textit{Rank-1}, \textit{mAP}, and \textit{mINP}. 
For CCPG, we report \textit{Rank-1} as the primary identification metric, together with the benchmark-defined summary results on the adopted evaluation subsets. 
For SUSTech1K, we report \textit{Rank-1} and \textit{Rank-5} under the official cross-view protocol, and additionally report attribute-wise Rank-1 results when using the benchmark-defined probe subsets. 
All metrics are computed using the student model only at inference.

\subsection{Computational Profiles and Gap-Closing Analysis}
\label{sec:exp_eff}

Fig.~\ref{fig:f3} compares representative gait recognition models in terms of backbone FLOPs, parameter count, and Rank-1 accuracy on Gait3D~\cite{zheng2022gait}, where bubble size indicates the number of parameters.

Lightweight models such as GaitPart~\cite{fan2020gaitpart} and GaitGL~\cite{lin2022gaitgl} operate at relatively low computational cost but achieve limited accuracy. 
GaitBase~\cite{fan2023opengait} offers a more balanced trade-off, while larger models, including DeepGaitV2~\cite{fan2025opengait}, SkeletonGait++~\cite{fan2024skeletongait}, and SwinGait~\cite{fan2025opengait}, deliver higher accuracy at the expense of substantially increased FLOPs. 
These observations motivate our teacher-student setting: we take GaitBase as a representative lightweight deployment model and leverage higher-capacity architectures as teachers to narrow the accuracy gap without increasing inference cost. 
As shown in Fig.~\ref{fig:f3}, GaitKD consistently improves the lightweight student while preserving its deployment efficiency.

\begin{figure}[t]
  \centering
  \includegraphics[scale=0.48]{compute_profile_scatter.png}
  \caption{\textbf{Computational profiles of different gait recognition models.}
Backbone FLOPs versus rank-1 accuracy on Gait3D, with bubble size indicating parameter count.}
  \label{fig:f3}
  \vspace{-1em}
\end{figure}

To further quantify the effect of distillation, Table~\ref{tab:gap_closing} reports how much of the teacher--baseline Rank-1 gap is recovered by the distilled student. 
For a teacher with Rank-1 $R_t$, baseline Rank-1 $R_b$, and distilled student Rank-1 $R_s$, we define
\begin{equation}
\text{GapClosed}=\frac{R_s-R_b}{R_t-R_b}\times 100\%.
\end{equation}
This metric is used only as an auxiliary analysis of distillation effectiveness; all primary results are still reported using the official benchmark metrics.

\begin{table}[t]
\centering
\caption{Gap-closing analysis on Gait3D (Rank-1). 
`Gain' denotes the improvement over the GaitBase baseline, and `GapClosed' measures the fraction of the teacher--baseline gap recovered by the distilled student. 
Teacher-only rows are reported for reference and do not have `Gain' or `GapClosed'.}
\label{tab:gap_closing}
\setlength{\tabcolsep}{3pt}
\renewcommand{\arraystretch}{1.1}
\scalebox{0.96}{
\begin{tabular}{c c c c c c}
\toprule
Setting & Teacher(s) & Student & Rank-1(\%) & Gain & GapClosed \\
\midrule
Baseline  & -- & GaitBase & 61.5 & -- & -- \\
Teacher   & DeepGaitV2 & -- & 74.4 & -- & -- \\
Distilled & DeepGaitV2 & GaitBase & 63.3 & +1.8 & 14.0\% \\
Distilled & DeepGaitV2 + SwinGait & GaitBase & 65.8 & +4.3 & 33.3\% \\
\bottomrule
\end{tabular}
}
\vspace{-1em}
\end{table}

\begin{table}[t]
\centering
\caption{Single-teacher distillation results on Gait3D.}
\label{tab:single_teacher_gait3d}
\setlength{\tabcolsep}{4pt}
\renewcommand{\arraystretch}{1.1}
\begin{tabular}{l l ccc}
\toprule
Student & Teacher & Rank-1(\%) & mAP(\%) & mINP(\%) \\
\midrule
GaitBase & -- & 61.5 & 52.3 & 34.4 \\
-- & SkeletonGait++ & 77.6 & 70.3 & 51.1 \\
-- & DeepGaitV2 & 74.4 & 65.8 & 46.4 \\
-- & SwinGait & 75.0 & 67.2 & -- \\
\midrule
GaitBase & SkeletonGait++ & 62.6 & 53.4 & 35.4 \\
GaitBase & DeepGaitV2 & 63.3 & 53.5 & 35.4 \\
GaitBase & SwinGait & 62.4 & 53.1 & 35.0 \\
\bottomrule
\end{tabular}
\vspace{-1em}
\end{table}

\begin{table}[t]
\centering
\caption{Single-teacher distillation results on CCPG. }
\label{tab:single_teacher_ccpg}
\setlength{\tabcolsep}{5pt}
\renewcommand{\arraystretch}{1.1}
\begin{tabular}{l l ccccc}
\toprule
Student & Teacher & CL(\%) & UP(\%) & DN(\%) & BG(\%) & Mean(\%) \\
\midrule
GaitBase & -- & 83.4 & 90.3 & 86.5 & 92.5 & 88.2 \\
-- & DeepGaitV2 & 90.5 & 96.3 & 95.3 & 96.7 & 94.7 \\
-- & SwinGait & 83.6 & 92.1 & 86.0 & 95.7 & 89.4 \\
-- & SkeletonGait++ & 90.1 & 95.4 & 92.5 & 97.3 & 93.8 \\
\midrule
GaitBase & DeepGaitV2 & 88.1 & 92.3 & 93.0 & 94.1 & 91.9 \\
GaitBase & SkeletonGait++ & 88.3 & 92.3 & 92.5 & 93.5 & 91.7 \\
GaitBase & SwinGait & 83.7 & 91.6 & 86.0 & 93.2 & 88.8 \\
\bottomrule
\end{tabular}
\vspace{-1em}
\end{table}

\begin{table}[t]
\centering
\caption{Single-teacher distillation results on SUSTech1K. }
\label{tab:single_teacher_sustech}
\setlength{\tabcolsep}{5pt}
\renewcommand{\arraystretch}{1.1}
\begin{tabular}{l l cc}
\toprule
Student & Teacher & Rank-1(\%) & Rank-5(\%) \\
\midrule
GaitBase & -- & 77.0 & 90.0 \\
-- & DeepGaitV2 & 82.3 & 92.5 \\
-- & SwinGait & 79.7 & 91.7 \\
-- & SkeletonGait++ & 81.3 & 95.5 \\
\midrule
GaitBase & SkeletonGait++ & 77.0 & 90.6 \\
GaitBase & DeepGaitV2 & 77.9 & 91.3 \\
GaitBase & SwinGait & 78.1 & 91.2 \\
\bottomrule
\end{tabular}
\vspace{-1em}
\end{table}

\begin{table}[t] 
\centering
\caption{
Multi-teacher distillation results across three benchmarks (Student: GaitBase). 
$\Delta_{\text{base}}$ is computed against the baseline; 
$\Delta_{\text{best}}$ is computed against the best single-teacher result.
}
\label{tab:multi_teacher_all}

\vspace{0.1em}
\textbf{(a) Results on Gait3D} \\
\vspace{0.2em}
\setlength{\tabcolsep}{3pt}
\renewcommand{\arraystretch}{1.1}
\resizebox{\columnwidth}{!}{
\begin{tabular}{l l c c c c c}
\toprule
Teacher Pair & Method & Rank-1(\%) & mAP(\%) & mINP(\%) & $\Delta_{\text{base}}$ & $\Delta_{\text{best-single}}$ \\
\midrule
\multirow{2}{*}{\shortstack[l]{SkeletonGait++\\ + DeepGaitV2}} 
& Mean-Teacher  & 63.4 & 53.6 & 35.8 & +1.9 & +0.1 \\
& GaitKD (Ours) & \textbf{64.0} & \textbf{54.4} & \textbf{36.4} & \textbf{+2.5} & \textbf{+0.7} \\
\midrule
\multirow{2}{*}{\shortstack[l]{DeepGaitV2\\ + SwinGait}} 
& Mean-Teacher  & 63.7 & 54.1 & 35.9 & +2.2 & +0.4 \\
& GaitKD (Ours) & \textbf{65.8} & \textbf{55.3} & \textbf{37.1} & \textbf{+4.3} & \textbf{+2.5} \\
\bottomrule
\end{tabular}
}

\vspace{1em} 
\textbf{(b) Results on CCPG} \\
\vspace{0.2em}
\resizebox{\columnwidth}{!}{
\begin{tabular}{l l c c c c c c c}
\toprule
Teacher Pair & Method & CL(\%) & UP(\%) & DN(\%) & BG(\%) & Mean(\%) & $\Delta_{\text{base}}$ & $\Delta_{\text{best-single}}$\\
\midrule
\multirow{2}{*}{\shortstack[l]{SkeletonGait++\\ + DeepGaitV2}} 
& Mean-Teacher  & 89.0 & 93.1 & 93.4 & 94.7 & 92.6 & +4.4 & +0.7 \\
& GaitKD (Ours) & \textbf{89.6} & \textbf{93.6} & \textbf{94.0} & \textbf{95.1} & \textbf{93.1} & \textbf{+4.9} & \textbf{+1.2} \\
\midrule
\multirow{2}{*}{\shortstack[l]{DeepGaitV2\\ + SwinGait}} 
& Mean-Teacher  & 88.7 & 92.5 & 93.2 & 94.1 & 92.1 & +3.9 & +0.2 \\
& GaitKD (Ours) & \textbf{88.9} & \textbf{92.7} & \textbf{93.6} & \textbf{94.3} & \textbf{92.4} & \textbf{+4.2} & \textbf{+0.5} \\
\bottomrule
\end{tabular}
}

\vspace{1em}
\textbf{(c) Results on SUSTech1K} \\
\vspace{0.2em}
\resizebox{\columnwidth}{!}{
\begin{tabular}{l l c c c c}
\toprule
Teacher Pair & Method & Rank-1(\%) & Rank-5(\%) & $\Delta_{\text{base}}$ & $\Delta_{\text{best-single}}$ \\
\midrule
\multirow{2}{*}{\shortstack[l]{SkeletonGait++\\ + DeepGaitV2}} 
& Mean-Teacher  & 78.0 & 91.1 & +1.0 & +0.1 \\
& GaitKD (Ours) & \textbf{78.2} & \textbf{91.4} & \textbf{+1.2} & \textbf{+0.3} \\
\midrule
\multirow{2}{*}{\shortstack[l]{DeepGaitV2\\ + SwinGait}} 
& Mean-Teacher  & 78.3 & 91.5 & +1.3 & +0.2 \\
& GaitKD (Ours) & \textbf{78.6} & \textbf{91.9} & \textbf{+1.6} & \textbf{+0.5} \\
\bottomrule
\end{tabular}
}
\end{table}

\subsection{Single-Teacher Distillation}
\label{sec:exp_single}

We first evaluate GaitKD under single-teacher distillation on three benchmarks: Gait3D, CCPG, and SUSTech1K. 
For completeness, we report both the standalone teacher performance (for reference) and the distilled student performance. 
All distilled results are obtained using \emph{student-only inference}. As shown in Tables~\ref{tab:single_teacher_gait3d}--\ref{tab:single_teacher_sustech}, GaitKD consistently improves the GaitBase student under single-teacher distillation across multiple benchmarks.

On Gait3D, all tested teachers improve the lightweight student over the GaitBase baseline. In particular, distillation from DeepGaitV2 yields the strongest gain, improving Rank-1 from 61.5\% to 63.3\%, while SkeletonGait++ and SwinGait also provide consistent improvements. This suggests that different high-capacity teachers can all transfer useful gait knowledge to the lightweight student, with the temporal modeling capacity of DeepGaitV2 being particularly beneficial in this setting. 

On CCPG, a similar trend can be observed under the cloth-changing protocol. All three teachers improve the student in terms of the overall mean accuracy, and the gains are especially clear for DeepGaitV2 and SkeletonGait++. Compared with the GaitBase baseline, distillation from DeepGaitV2 improves the mean score from 88.2\%  to 91.9\%, while SkeletonGait++ achieves a comparable improvement to 91.7\%. These results indicate that GaitKD can effectively transfer robustness from high-capacity teachers under challenging appearance changes. 

On SUSTech1K, the improvements are more modest but still consistent overall. DeepGaitV2 and SwinGait both improve the student in Rank-1 and Rank-5, while SkeletonGait++ mainly improves Rank-5 and leaves Rank-1 unchanged. Notably, SwinGait provides the best single-teacher Rank-1 result on this benchmark, suggesting that the most effective teacher may vary across sensing modalities and evaluation protocols. These results show that GaitKD is not tied to a specific teacher family. Instead, it consistently improves a lightweight gait baseline across diverse teacher architectures and benchmarks, while the relative benefit of each teacher depends on the dataset characteristics.

\subsection{Multi-Teacher Distillation: Complementary Supervision}
\label{sec:exp_multi}
We further evaluate GaitKD in a multi-teacher setting to examine whether heterogeneous teachers provide complementary supervision. 
As a reference, we introduce a simple baseline, denoted as \emph{Mean-Teacher}, which averages the distillation losses from individual teachers.

As shown in Tables~\ref{tab:multi_teacher_all}, multi-teacher supervision is beneficial across all three benchmarks. 
On Gait3D, Mean-Teacher already improves upon the corresponding best single-teacher result, suggesting that different teachers provide complementary supervision. 
Nevertheless, GaitKD yields further gains under the same teacher combinations, indicating that effective multi-teacher distillation requires more than direct averaging of teacher signals. 
In particular, the combination of \emph{DeepGaitV2 + SwinGait} achieves the best overall result on Gait3D, reaching 65.8\% Rank-1, 55.3\% mAP, and 37.1\% mINP.

A similar trend can be observed on CCPG. 
For both teacher pairs, multi-teacher distillation further improves the lightweight student over the best constituent single-teacher setting, while GaitKD consistently performs better than Mean-Teacher. 
In particular, the combination of \emph{SkeletonGait++ + DeepGaitV2} yields the best overall mean score of 93.1\%.

On SUSTech1K, the gains are relatively smaller but remain consistent. 
Both teacher combinations improve the lightweight student over their corresponding single-teacher baselines, and GaitKD again achieves the best overall performance. 
The strongest result is obtained by \emph{DeepGaitV2 + SwinGait}, which reaches 78.6\% Rank-1 and 91.9\% Rank-5.

These results suggest that the benefit of multi-teacher distillation is not merely due to more supervision, but also depends on how complementary teacher knowledge is integrated. 
By explicitly decoupling decision-level transfer and boundary-level transfer, GaitKD provides a more effective way to aggregate heterogeneous teachers than naive loss averaging.

\subsection{Homogeneous Distillation: Same-Family Teacher--Student Transfer}
\label{sec:exp_homo}
To further examine whether GaitKD is also effective beyond heterogeneous teacher--student pairs, we evaluate a homogeneous distillation setting using the same backbone family with different capacities. Specifically, we use a lightweight \emph{DeepGaitV2-14} model as the student and a stronger \emph{DeepGaitV2-22} model as the teacher. This setting removes cross-architecture mismatch and allows us to test whether the proposed decision-level and boundary-level transfers remain beneficial within the same model family.

Table~\ref{tab:homo_distill} reports the results. Compared with the student baseline, GaitKD consistently improves all three metrics under homogeneous distillation, showing that the proposed framework is not restricted to heterogeneous teacher--student settings. Although the improvement is smaller than the teacher--student gap, the distilled student recovers part of the performance of the deeper teacher while preserving the efficiency of the shallower deployment model. These results suggest that GaitKD provides a generally effective distillation mechanism for gait recognition, while being particularly useful in more challenging heterogeneous settings.

\begin{table}[t]
\centering
\caption{Homogeneous distillation results using the same backbone family. The student is DeepGaitV2-14 and the teacher is DeepGaitV2-22.}
\label{tab:homo_distill}
\setlength{\tabcolsep}{5pt}
\renewcommand{\arraystretch}{1.1}
\begin{tabular}{llccc}
\toprule
Teacher & Student & Rank-1(\%) & mAP(\%) & mINP(\%) \\
\midrule
-- & DeepGaitV2-14 & 72.3 & 64.0 & 43.4 \\
DeepGaitV2-22 & -- & 74.4 & 65.8 & 46.4 \\
DeepGaitV2-22 & DeepGaitV2-14 & 73.2 & 64.9 & 44.8 \\
\bottomrule
\end{tabular}
\vspace{-1em}
\end{table}

\subsection{Ablation Study: Loss Design and Alternatives}
\label{sec:exp_ablation}

With the teacher-student pair fixed (DeepGaitV2 $\rightarrow$ GaitBase), we vary the distillation objective to isolate each component’s effect. Results are shown in Table~\ref{tab:loss_ablation}.

\paragraph{Decision-level vs.\ boundary-level transfer}
Decision-level distillation alone improves the baseline, indicating that transferring the teacher's softened class relations is effective.
Boundary-level distillation alone yields limited gains, but becomes more effective when coupled with decision transfer.
The full GaitKD objective achieves the best Rank-1 among tested variants, validating the complementarity between
part-calibrated decision transfer and boundary-preserving representation regularization.

\paragraph{Comparison to feature regression and alternative KD objectives}
We compare boundary transfer to point-wise feature regression (MSE) and evaluate an alternative logit KD objective (Naive KD~\cite{hinton2015distilling}, KALSR~\cite{wen2021preparing}, and KAPS~\cite{wen2021preparing}).
Different objectives trade off Rank-1, mAP, and mINP differently; overall, our method provides the strongest Rank-1,
while alternatives may improve specific retrieval summaries.
These results support the design choice of boundary-preserving distillation under heterogeneous teacher-student mismatch.

\begin{table}[t]
\centering
\caption{Ablation on distillation objectives (teacher: DeepGaitV2, student: GaitBase).}
\label{tab:loss_ablation}
\begin{tabular}{l l ccc}
\toprule
Logits-level & Feature-level & Rank-1(\%) & mAP(\%) & mINP(\%) \\
\midrule
Naive KD & \textbf{\checkmark} & 62.0 & 52.7 & 34.8 \\
KALSR & \textbf{\checkmark} & 60.1 & 50.1 & 32.3 \\
KAPS & \textbf{\checkmark} & 61.8 & 53.0 & 35.0 \\

\midrule
MSE & \textbf{\checkmark} & 61.5 & 51.6 & 33.8 \\
\textbf{$\times$} & MSE & 62.7 & 53.6 & 35.1 \\
MSE & MSE & 62.6 & 53.0 & 35.3 \\
\textbf{\checkmark} & MSE & 62.4 & 53.4 & 35.7 \\
\midrule
\textbf{$\times$} & \textbf{$\times$} & 61.5 & 52.3 & 34.4 \\
\textbf{\checkmark} & \textbf{$\times$} & 62.5 & 52.9 & 35.3 \\
\textbf{$\times$} & \textbf{\checkmark} & 61.8 & 52.0 & 33.6 \\
\textbf{\checkmark(Ours)} & \textbf{\checkmark(Ours)} & 63.3 & 53.5 & 35.4\\
\bottomrule
\end{tabular}
\vspace{-1em}
\end{table}

The experiments form a consistent evidence chain for GaitKD.
We quantify the efficiency-accuracy gap that motivates distillation, show consistent gains under heterogeneous single-teacher distillation,
demonstrate additional benefits from multi-teacher supervision, and validate the complementary roles of decision-level and boundary-level transfers.

\subsection{Feature Visualization}
\label{sec:exp_vis}

\begin{figure}[t]
    \centering
    \includegraphics[width=0.48\textwidth]{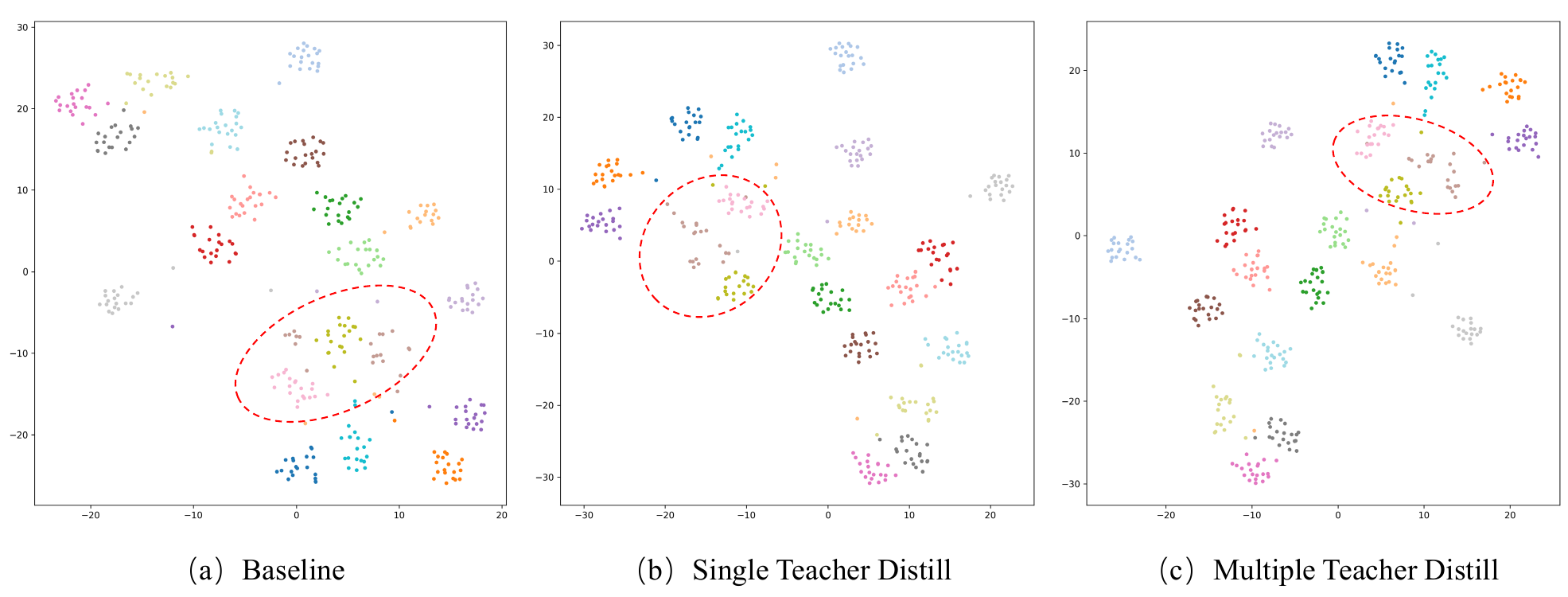}
    \caption{\textbf{Feature visualization of the student embeddings under different training settings.}
    (a) Baseline student without distillation.
    (b) Student trained with single-teacher distillation.
    (c) Student trained with multi-teacher distillation.
    The dashed red circles highlight a challenging region in the feature space.
    Compared with the baseline, distillation leads to more compact intra-class clustering and clearer inter-class separation, while the multi-teacher setting further improves the organization of ambiguous samples.}
    \label{fig:vis_feat}
    \vspace{-1em}
\end{figure}

To provide a more intuitive understanding of the learned representation space, we visualize the student embeddings under three settings: 
(a) the baseline student trained without distillation, 
(b) the student distilled from a single teacher (DeepGaitV2), and 
(c) the student distilled from multiple teachers (DeepGaitV2 + SwinGait). 
Fig.~\ref{fig:vis_feat} shows the corresponding two-dimensional feature distributions.

Compared with the baseline, single-teacher distillation already produces a cleaner embedding space: samples from the same identity become more compact, and several nearby identity clusters become better separated. 
This suggests that the proposed decision-level and boundary-level transfer can effectively regularize the student representation beyond the standard task loss.
The multi-teacher setting further improves the feature organization. 
As highlighted by the dashed red circles in Fig.~\ref{fig:vis_feat}, the ambiguous region that exhibits noticeable overlap in the baseline becomes progressively more structured after distillation, and the separation between neighboring identities is more clearly preserved. 
This observation is consistent with the quantitative results in the previous subsections: multi-teacher supervision provides complementary knowledge, and GaitKD makes better use of this supervision by jointly transferring decision relations and embedding-space boundaries.

The visualization indicates that GaitKD improves the discriminability of the student representation, leading to tighter intra-class clustering and clearer inter-class separation, especially in challenging regions where the baseline features are less well organized.

\section{Conclusion}
\label{sec:conclusion}

We have studied knowledge distillation for gait recognition under heterogeneous teacher-student configurations.
We have introduced GaitKD, an effective distillation framework that explicitly decouples the transfer of
\emph{decision geometry} and \emph{representation boundaries} through a unified part-wise interface. GaitKD integrates part-calibrated logit distillation via temperature-scaled KL divergence with
boundary-level embedding distillation via activation-boundary constraints, enabling robust transfer without architectural modification
or inference-time overhead. Extensive experiments demonstrate that GaitKD consistently improves a lightweight gait baseline when distilled from
diverse high-capacity teachers, and that multi-teacher supervision provides additional gains, supporting the complementarity of
heterogeneous knowledge sources.
Ablation studies further confirm the distinct roles of decision-level and boundary-level transfers and highlight the advantage of
boundary-preserving distillation over point-wise feature regression under teacher-student mismatch.
Overall, GaitKD provides a practical and effective recipe for narrowing the gap between high-capacity training-time teachers and efficient deployment-time gait models.

{
\bibliographystyle{ieeetr}
\bibliography{egbib}
}
\vspace{-1.7cm}

\end{document}